\documentclass[runningheads]{llncs}
\usepackage[T1]{fontenc}
\usepackage{graphicx}
\usepackage{times}
\usepackage{latexsym}
\usepackage{hyperref}
\usepackage{bbding}
\usepackage{amsmath,amssymb}

\begin{document}
\title{IIU: Independent Inference Units for Knowledge-based Visual Question Answering}
%
%
\author{Yili Li \inst{1,2} \and
Jing Yu\inst{1,2}\inst{(}\Envelope\inst{)} \and
Keke Gai\inst{3} \and
Gang Xiong\inst{1}}
\institute{Institute of Information Engineering, Chinese Academy of Sciences, Beijing, China 
\email{\{liyili, yujing02, xionggang\}@iie.ac.cn}
\and
School of Cyber Security, University of Chinese Academy of Sciences, Beijing, China \and
School of Cyberspace Science and Technology, 
Beijing Institute of Technology, Beijing, China \\
\email{gaikeke@bit.edu.cn}}

%
\maketitle              
\begin{abstract}
Knowledge-based visual question answering requires external knowledge beyond visible content to answer the question correctly. One limitation of existing methods is that they focus more on modeling the inter-modal and intra-modal correlations, which entangles complex multimodal clues by implicit embeddings and lacks interpretability and generalization ability. The key challenge to solve the above problem is to separate the information and process it separately at the functional level. By reusing each processing unit, the generalization ability of the model to deal with different data can be increased. In this paper, we propose \textbf{I}ndependent \textbf{I}nference \textbf{U}nits (\textbf{IIU}) for fine-grained multi-modal reasoning to decompose intra-modal information by the functionally independent units. Specifically, IIU processes each semantic-specific intra-modal clue by an independent inference unit, which also collects complementary information by communication from different units. To further reduce the impact of redundant information, we propose a memory update module to maintain semantic-relevant memory along with the reasoning process gradually. In comparison with existing non-pretrained multi-modal reasoning models on standard datasets, our model achieves a new state-of-the-art, enhancing performance by 3\%, and surpassing basic pretrained multi-modal models. The experimental results show that our IIU model is effective in disentangling intra-modal clues as well as reasoning units to provide explainable reasoning evidence. Our code is available at https://github.com/Lilidamowang/IIU.

\end{abstract}

\section{Introduction}
The knowledge-based visual question answering task requires agent to answer questions according to visual content and external knowledge. Compared to AI agent, human can easily combine visual observation with external knowledge to answer questions. Recent development on multi-modal inference mainly focuses on incorporating multimodal information in the reasoning process. \cite{narasimhan2018out} uses GCN to reason on fact graphs with image-question-entity embeddings. Mucko \cite{zhu2020mucko} utilizes a heterogeneous graph depicting images with layers of visual, semantic, and knowledge-based information. Following heterogeneous graphs reasoning \cite{zhu2020mucko}, \cite{yu2020cross} introduces a parallel reasoning model for separate modal processing, enhancing generalization and extendibility. 

\begin{figure*}[t]
\vspace{-5pt}
    \centering
    \includegraphics[width=\textwidth]{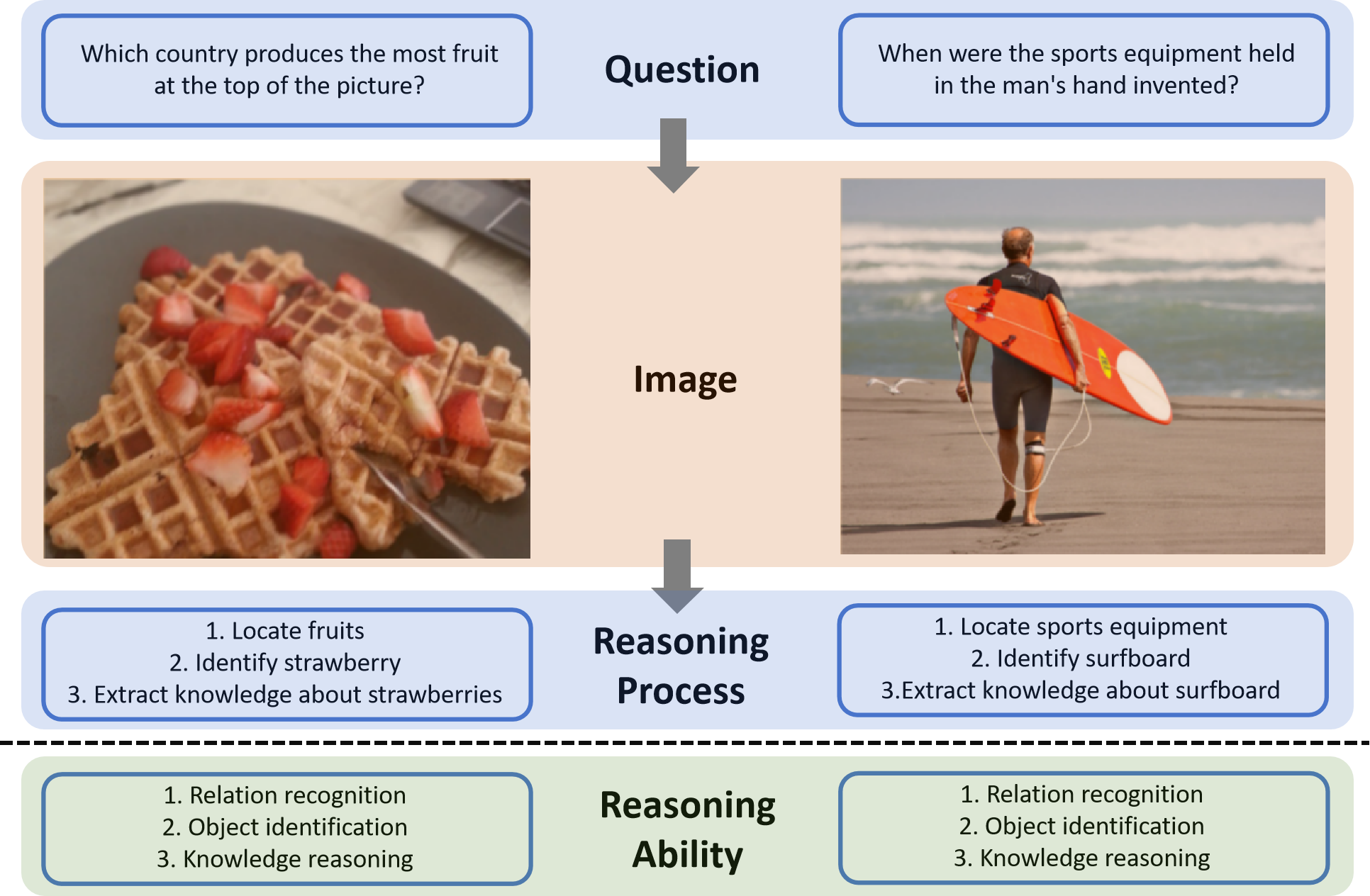}
    \caption{Illustration of our motivation. The question on the left and right have different processing objects, but they require the same reasoning abilities. By combining and reusing these reasoning abilities, it is possible to use the same reasoning process to answer different questions.}
    \label{fig:model_mov}
    \vspace{-10pt}
\end{figure*}

However, these methods generally process the question-relevant information through a single reasoning module or gather question-relevant information  from different modalities independently. A more fine-grained approach, processing only the relevant parts of information differently, can simplify training and boost the model's generalization capability across diverse datasets. As shown in Figure \ref{fig:model_mov}, for two different questions and images, they have different reasoning processes. For example, for the question on the left, the model needs to first identify the located fruit as a strawberry, and then based on relevant knowledge about strawberries, provide the name of the country with the highest production as the answer. In contrast, for the question on the right, the model needs to locate and identify the surfboard in the man's hand, and then answer the question based on relevant knowledge. Although the two questions have different processing objects, the required reasoning abilities for an AI agent are similar between the two questions, including relation recognition, object identification, and external knowledge reasoning. By combining and reusing different abilities, reasoning for more questions can be achieved, improving the model's accuracy and generalization.

In this paper, we propose \textbf{I}ndependent \textbf{I}nference \textbf{U}nits  (IIU) for fine-grained multi-modal reasoning to decompose intra-modal information by the functionally independent units. Specifically, we utilize the multi-modal heterogeneous graph to extract question-relevant information as memory for reasoning from the original information. Then, we define the function of each reasoning unite by an independent GRU module, and each reasoning step activates a specific number of units through the attention mechanism. The information is decomposed through independent units at the functional level. Besides, to model the reasoning chains between independent inference units, we utilize units communication to organize the question-relevant inference units and  pass internal reasoning results according to the reasoning sequence. To ensure that the semantic information is not distorted in the reasoning process and further reduce the impact of redundant information, we propose a memory reduction module to maintain semantic information in the process of reasoning. 

The main contributions of this paper are summarized as follows: (1) We decompose multi-modal information functionally for modular reasoning, enhancing the model's generalization ability by independent inference units with specific reasoning abilities. (2) By separating information at the functional level, we effectively process redundant information, thus bolstering model stability. (3) We visualize the units activated when processing different information of different modalities, and demonstrate good interpretability of our approach. 
\section{Related Work}

\noindent \textbf{Visual Question Answering.} Visual Question Answering (VQA) task requires agent to answer questions according to visual content, which needs the ability to understand and process multi-modal information. 
In order to reduce the impact of redundant information, \cite{xu2016ask} \cite{anderson2018bottom} apply the attention mechanism to VQA tasks from different perspectives.
A trend to solve VQA task is to represent information through graph structure and complete reasoning on the graph. \cite{wang2019neighbourhood} and \cite{li2019relation} describe the objects and relationships in the image by constructing a scene graph. 
This method is also effective for other visual tasks. 
In knowledge-based visual question answering tasks, instead of building a scene graph, \cite{narasimhan2018out} build a fact graph and each node is represented by the fixed form of image-question-entity embedding. However, the visual information is wholly provided which may introduce redundant information for prediction. To reduce noise and combine information from different modalities, \cite{zhu2020mucko} depict an image by multi-layer graphs and perform cross-modal heterogeneous graph reasoning on them to capture complementary evidence from different layers that are most relevant to the question. However, reasoning on graph still focuses on modeling the inter-modal and intra-modal relationships in the Visual Question Answering task. Lack of more fine-grained reasoning methods to extract complex clues within the modal and increase the interpretability.

\noindent \textbf{Separate Recurrent Models.}
In multi-step reasoning for VQA tasks, recurrent models like GRU are commonly used. However, our objective is to implement an independent reasoning mechanism. IndRNN \cite{li2018independently}, a distinct separate recurrent model where each unit operates independently, lays the foundation for this approach. Building on this, RIM \cite{goyal2019recurrent} introduces sparse inter-unit communication via an attention mechanism. Drawing inspiration from RIM, our model's inference units not only reason independently but also gather complementary information.

\begin{figure*}[ht]
\vspace{-5pt}
    \centering
    \includegraphics[width=\textwidth]{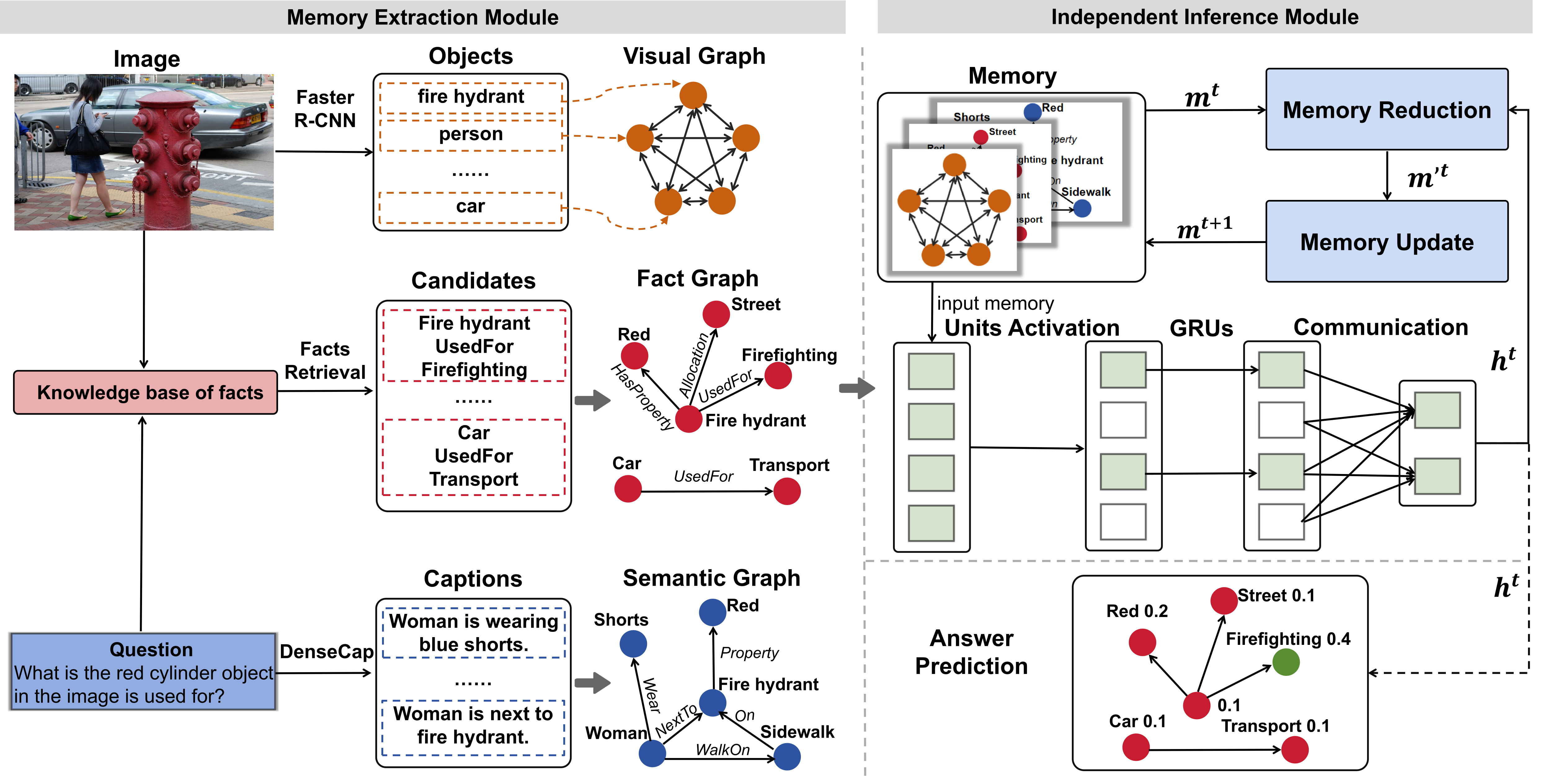}
    \caption{An overview of our model. The model contains two main modules: Memory Extraction Module, Independent Inference Module. After $t$ steps of inference, we utilize hidden states of each units to achieve answer prediction on fact graph.}
    \label{fig:model_overview}
    \vspace{-10pt}
\end{figure*}
\section{Methodology}
The task aims to predict an answer while reasoning uses external knowledge to construct memory graphs. The structure of our model is shown in Figure \ref{fig:model_overview}. We depict an image by three layers of graphs, including visual graph, semantic graph, and fact graph respectively, imitating the understanding of various properties of an object and relationships. Then we perform independent reasoning based on memory graph by 
Independent Inference Module which consists of two parts. By stacking the two processes multiple times, our model realizes the reasoning of disentangling information from the functional level and reduces the influence of redundant information.

\subsection{Multi-Modal Graph Construction}
As shown in Figure \ref{fig:model_overview}, we utilize Multi-modal Heterogeneous Graph which is proposed in \cite{zhu2020mucko} to represent the question-image pairs. Specifically, the nodes in the fully-connected visual graph are objects extracted by Faster R-CNN, and edges are the relative spatial relationships between two objects. In addition to visual information, we use dense captions to extract a set of local-level semantics in an image. Then utilize semantic graph parsing model \cite{anderson2016spice} to construct Semantic Graph, where nodes are the name or attribute and edges are relationships. To find the optimal supporting-fact, we retrieve relevant candidate facts from the knowledge base following a score-based approach \cite{narasimhan2018out}. Each node in Fact Graph denotes an entity in the supporting-fact set and is represented by GloVe embedding of the entity, and each edge denotes the relationship between two entities.

\subsection{Independent Units Inference}
Independent Units Inference is our main module. The key is to reason independently by dividing memory from functional level and obtaining complementary information from each independent part. The Independent Units Inference consists of three parts: \emph{Units Activation }, \emph{Internal Dynamics} and \emph{Communication}.

\noindent \textbf{Units Activation.}
This module learns to dynamically select the unit most relevant to the input. The default hidden state of each unit is set as question embedding after the last state of LSTM. Then each reasoning step, we active the top-$k_a$ units through the attention score of hidden state and input. First we concatenate a row of zeros for current input $m^{(t+1)}$: $M = \emptyset \oplus m^{(t+1)}$. Then, we use linear transformation to construct keys, values and queries($K = MW^e$, $ V = MW^v $, $ Q = h^{(t)}W^q_k $ ). The soft-attention value of each unit is expressed as $A^{in}_k$.

\noindent \textbf{Internal Dynamics of Units.}
Each activated unit needs to process the input memory independently and output the new inference state.
We utilize Gate Recurrent Unit(GRU) \cite{chung2015recurrent} to process the input memory and generate the intermediate state $\hat{h}^{(t)}$ of units. We refer to the activated set as $\mathcal{S}_t$. The operation of each activated unit is as follows,
\begin{eqnarray}
\hat{h}^{(t)}_k = GRU(h^{(t)}_k, A^{in}_k) , \forall k \in \mathcal{S}_t
\end{eqnarray}

\noindent \textbf{Communication.}
Although each unit works independently, we still need to obtain complementary information between independent parts. We update the inference state by following attention-like methods,
\begin{eqnarray}
h_k^{(t+1)} = softmax(\frac{Q^{(t)}_k(K^{(t)})^T}{\sqrt{d}})V^{(t)} + \hat{h}^{(t)}_k, \forall k \in \mathcal{S}_t
\end{eqnarray}
where $Q^{(t)}_k = \hat{W}^q_k\hat{h}^{(t)}_k, \forall k \in \mathcal{S}_t$, $K^{(t)}_k = \hat{W}_k^e\hat{h}^{(t)}_k, \forall k$ and $V^{(t)}_k=\hat{W}^v_k\hat{h}^{(t)}_k, \forall k$.
In this way, while realizing disentangling reasoning, it also retains the complementary information of each part.

\subsection{Question-guided Memory Update}
Before entering the next reasoning step, the model needs to maintain memory information based on the current inference state $h^{(t)}$. This module aims to prevent distortion of memory information and update the memory as indicated by the question status.

We first reduce memory to zero in each step of reasoning to reduce information that has been inferred. Each node of the memory graph concatenate a row full of zeros, which is represented as $\hat{v}_i^{(t)}$. Then, we use the attention mechanism between zeros and the original representation of memory to obtain results after reduction by the guide of the current inference state $h^{(t)}$. we get the attention weights as follows,
\begin{eqnarray}
\alpha_i^{R} = softmax(W_{\alpha}^{R}Tanh(W_{v}^{R}\hat{v}_i^{(t)} + W_{h}^{R}h^{(t)}))
\end{eqnarray}
where all $W$ are learned parameters. $h^{(t)}$ is the current inference state produced by last inference step. The representation of each node after reduction is computed,
\begin{eqnarray}
v_i^{(t)} = \alpha_i^{R} \cdot \hat{v}_i^{(t)}
\end{eqnarray}

Unlike the static memory representations, our proposed knowledge memory will be updated adaptively during each reasoning step. The relationship between nodes as follows,

\begin{eqnarray}
rel_i = \sum_{k \in \mathcal{N}_i}W_{rel}[v_k^{(t)}, r_{ki}]
\end{eqnarray}
where $\mathcal{N}_i$ represents a set of 1-hop neighboring nodes regarding the memory entity $v_i$. $[a, b]$ represents b is concatenated behind a. $r_{ki}$ is a edge from node $k$ to node $i$. Then we update the node representation with $h^{(t)}$ and the relationship $rel$,
\begin{eqnarray}
v_i^{(t+1)} = W_v^{rel}[v_i^{(t)},rel_i,h^{(t)}]
\end{eqnarray}

\subsection{Answer Prediction and Training}
To predict the answer, we update the representation $v_i^{F}$ with the last inference state $h^{(t)}$ via the gate mechanism,
\begin{eqnarray}
gate_{i} = \sigma(W_{gate}[h^{(t)}, v_i^{F}])
\end{eqnarray}
\begin{eqnarray}
\hat{v}_{i}^{F} = W_{rep}[gate_i \circ [h^{(t)}, v_i^{F}] ]
\end{eqnarray}
where $\sigma$ is sigmoid function, and `$\circ$' represents element-wise product, and ${v}_{i}^{F}$ represents the node in the fact graph.

All the concepts $\hat{v}_{i}^{F}$ are fed into a binary classification to predict the answers. Then we use weighted binary cross-entropy loss to deal with the imbalanced training data,
\begin{eqnarray}
l_n = - \sum [a \cdot y_i ln\hat{y}_i + b \cdot (1-y_i)ln(1-\hat{y}_i)]
\end{eqnarray}
where $y_i$ is the ground truth label for $v_i^{F}$ and a,b represent the weights for negative and positive samples.

\section{Experiment}
\textbf{Implementation Details}. We set a=0.7 and b=0.3 in the binary cross-entropy loss. Our model is trained by Adam optimizer with 30 epochs, where the batch size is 32. Warm up strategy is applied for 2 epochs with initial learning rate $1 \times 10^{-3}$ and warm-up factor 0.2. In Independent Units Inference, we set 8 units and active 4 units in each step. In this case, the model's parameter size is only 74M, with an average training time of 0.4 hours per epoch with a single V100-32GB GPU, ensuring lightweight training and deployment processes.

\noindent \textbf{Dataset}. We evaluate IIU on Outside Knowledge VQA(OK-VQA) dataset \cite{marino2019ok}, which contains 14,031 images and a total of 14,055 questions, covering 10 different categories, such as Vehicles and Transportation, Sports and Recreation, Cooking and Food, etc. In this work, we retrieve the supporting knowledge from ConceptNet. Besides, we alse evaluate on the FVQA \cite{wang2017fvqa} dataset, which consists of 2,190 images, 5,286 questions and a knowledge base of 193,449 facts. 

\subsection{Comparison with State-of-the-Art Methods}

\begin{table}
\begin{center}
\vspace{-15pt}
\caption{State-of-the-art comparison on the OK-VQA dataset} \label{tab:okvqa}
\setlength{\tabcolsep}{1.5mm}{
\begin{tabular}{c|c}
    \hline
    Method & overall accuracy  \\
    \hline
    Q-Only \cite{marino2019ok} & 14.93  \\
    MLP \cite{marino2019ok} & 20.67 \\
    BAN \cite{kim2018bilinear} & 25.17 \\
    MUTAN \cite{ben2017mutan} & 26.41  \\
    BAN + AN \cite{marino2019ok} & 25.61  \\
    MUTAN + AN \cite{marino2019ok} & 27.84  \\
    BAN/AN oracle \cite{marino2019ok} & 27.59  \\
    MUTAN/AN oracle \cite{marino2019ok} & 28.47  \\
    CBM (T5-Small) \cite{salaberria2023image} & 29.20\\
    GRUC \cite{yu2020cross} & 29.87 \\
    Mucko \cite{zhu2020mucko} & 29.20 \\
    KM$^{4}$ \cite{zheng2021km4} & 31.32 \\
    CBM (BERT) \cite{salaberria2023image} & 32.50\\
    \hline
    ViLBERT \cite{lu2019vilbert} & 31.35 \\
    LXMERT \cite{tan2019lxmert} & 32.04 \\
    ConceptBert \cite{garderes2020conceptbert} & 33.66 \\
    \hline
    \textbf{IIU (ours)} & \textbf{34.87}\\
    \hline
\end{tabular}}
\vspace{-15pt}
\end{center}
\end{table}

\begin{table}[h]
\begin{center}
\vspace{-15pt}
\caption{State-of-the-art comparison on FVQA dataset} \label{tab:fvqa}
\begin{tabular}{c|c}
    \hline
    Method & accuracy \\
    \hline
    LSTM-Question+Image+Pre-VQA & 24.98 \\
    Hie-Question+Image+Pre-VQA & 43.14 \\
    FVQA (top-3-QQmaping) \cite{wang2017fvqa} & 56.91 \\
    FVQA (Ensemble) \cite{wang2017fvqa} & 58.76 \\
    Straight to the Facts (STTF) \cite{narasimhan2018straight} & 62.20 \\
    Reading Comprehension \cite{li2019visual} & 62.96 \\
    Out of the Box (OB) \cite{narasimhan2018out} & 69.35 \\
    Mucko \cite{zhu2020mucko} & 73.06 \\
    GRUC \cite{yu2020cross} & 79.63 \\
    \hline
    Human & 77.99 \\
    \hline
    \textbf{IIU(ours)} & \textbf{86.37} \\
    \hline
\end{tabular}
\vspace{-15pt}
\end{center}
\end{table}

In Table \ref{tab:okvqa}, we compare our IIU model with other low-resource multi-modal inference model on the OK-VQA dataset, including the VQA models, knowledge-based VQA models, and ensemble models. The VQA models contain Q-Only \cite{marino2019ok}, MLP \cite{marino2019ok}, BAN \cite{kim2018bilinear}, MUTAN \cite{ben2017mutan}. The knowledge-based VQA models consist of BAN+AN, MUTAN+AN, and some multi-step reasoning model, such as GURC \cite{yu2020cross}, Mucko \cite{zhu2020mucko}, KM$^{4}$ \cite{zheng2021km4}. The ensemble models are BAN/AN oracle and MUTAN/AN oracle. Besides, we also compare with some basic pre-trained models in multi-modal learning, such as ViLBERT \cite{lu2019vilbert}, LXMERT \cite{tan2019lxmert} and ConceptBert \cite{garderes2020conceptbert}. We can see from the result in Table \ref{tab:okvqa}, our IIU model has the highest overall accuracy. Especially, our model achieves a 5\% boost on overall accuracy compared with GRUC and Mucko. We all get answers from reasoning on the fact graph, but Mucko and GRUC only decompose information from the modal level. By further disentangling the information at the functional level, the performance of our model has been greatly improved. Besides, our model outperforms some pre-trained models in overall performance. Besides, in Table \ref{tab:fvqa}, we also compare our IIU model on FVQA dataset with state-of-the-art models without pre-training.

\subsection{Ablation Study}
Our IIU model is mainly composed of two parts in Independent Inference Module, Question Guide Memory Update, and Independent Units Inference. The reasoning information is decomposed through the functionally independent units. The processing of redundant information requires the cooperation of two modules. In Table \ref{tab:ablation}, we first study the influence of each module on the optimal accuracy without limiting the reasoning steps. In Table \ref{tab:ablation} model `2', we turn Independent Units to one single GRU which removes the information disentangling, and the performance is 11\% lower than the full model. We can also see in model `3' that it is necessary to obtain complementary information from other units. Once we cut off the communication between the independent units, the results of the model dropped slightly by about 3\%.
We can see from model `1' that the Memory Update module will not have a great impact on the optimal accuracy, but still 2\% lower than the full model. Because this module can help the model maintain memory information, on one hand, it removes used or redundant memories, and on the other hand, it helps the model update memory representations based on the state of inference. Besides the structure of our model, we also study the influence of modal information in Table \ref{tab:ablation} model `5', `6' and `7'.


\begin{table}[t]
\begin{center}
\caption{Ablation study on OK-VQA dataset. } \label{tab:ablation}
\begin{tabular}{c|c|c}
    \hline
    \multicolumn{2}{c|}{\textbf{Ablation of module}} & \textbf{Accuracy} \\
    \multicolumn{2}{c|}{IIU (full model)} & \textbf{34.87}\\
    \hline
    1 & w/o Memory Update & 32.30\\
    2 & w/o Independent Units & 23.42\\
    3 & w/o Units Communication & 29.39 \\
    4 & \quad w/o Memory Update \& Independent Units \quad \quad & 22.37 \\
    \hline
    \multicolumn{2}{c|}{\textbf{Ablation of modalities}} & \textbf{Accuracy} \\
    \hline
    5 & w/o Semantic Graph & 32.20 \\
    6 & w/o Visual Graph & 33.17 \\
    7 & w/o Semantic Graph \& Visual Graph & 18.20 \\
    \hline
\end{tabular}
\vspace{-25pt}
\end{center}
\end{table}

\subsection{Interpretability and Visualization}

\begin{figure*}[h]
    \centering
    \includegraphics[width=\textwidth]{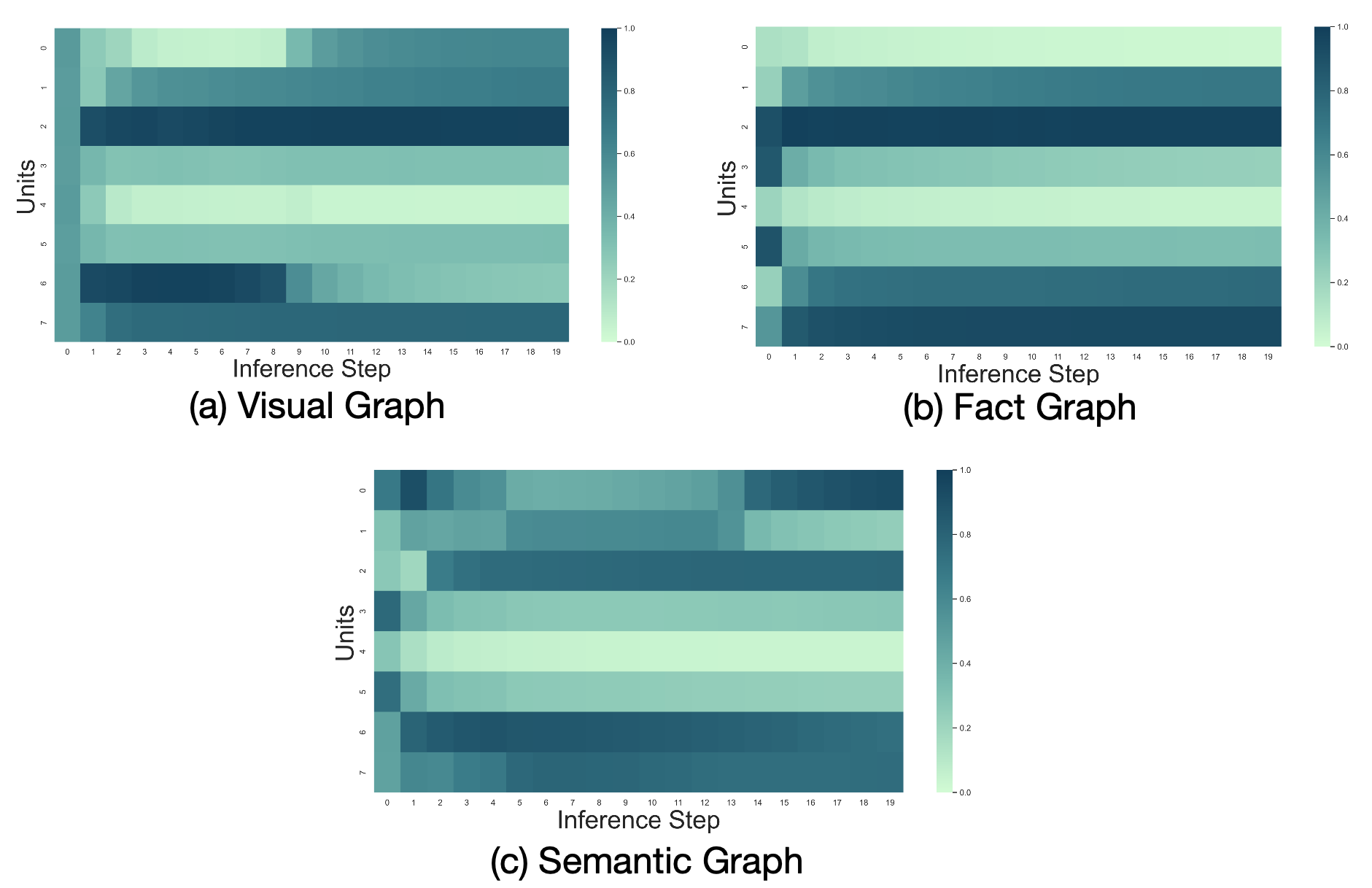}
    \caption{Units Activation of Different Modalities. Dark indicates active, and light indicates inactive.}
    \label{fig:different-modalities}
    \vspace{-15pt}
\end{figure*}

In this section we will show how units decompose information and focus on a specific functional level. We set up 8 units, and the inference step is 20. In Figure \ref{fig:different-modalities}, we show the average units activated by three different modalities information in each inference step. The information of these three modalities comes from Visual Graph, Semantic Graph, and Fact Graph respectively. We can see that the information of different modalities has different activation distributions of units. In modal of Visual Graph, the units of numbers 2, 6, and 7 are activated. Units No.1, 2 and 7 are activated in modal of Fact Graph, and 0, 1, 2, 6, and 7 are activated in modal of Semantic Graph. The activated units in the Semantic Graph modal contain other two modalities. The reason for this result is that Semantic Graph is a high-level representation of the image. So the reasoning process will overlap at the functional level with the modal of Visual Graph and activate the same units. Besides, the concepts in the Fact Graph are included in the Visual modal and the Semantic modal, accordingly the units for Fact modal reasoning may helpful to other modalities. At last, this experiment shows that IIU can decompose the input information and deal with each kind of information differently.

\subsection{Model Stability}
\begin{figure*}[h]
    \centering
    \includegraphics[width=0.7\textwidth]{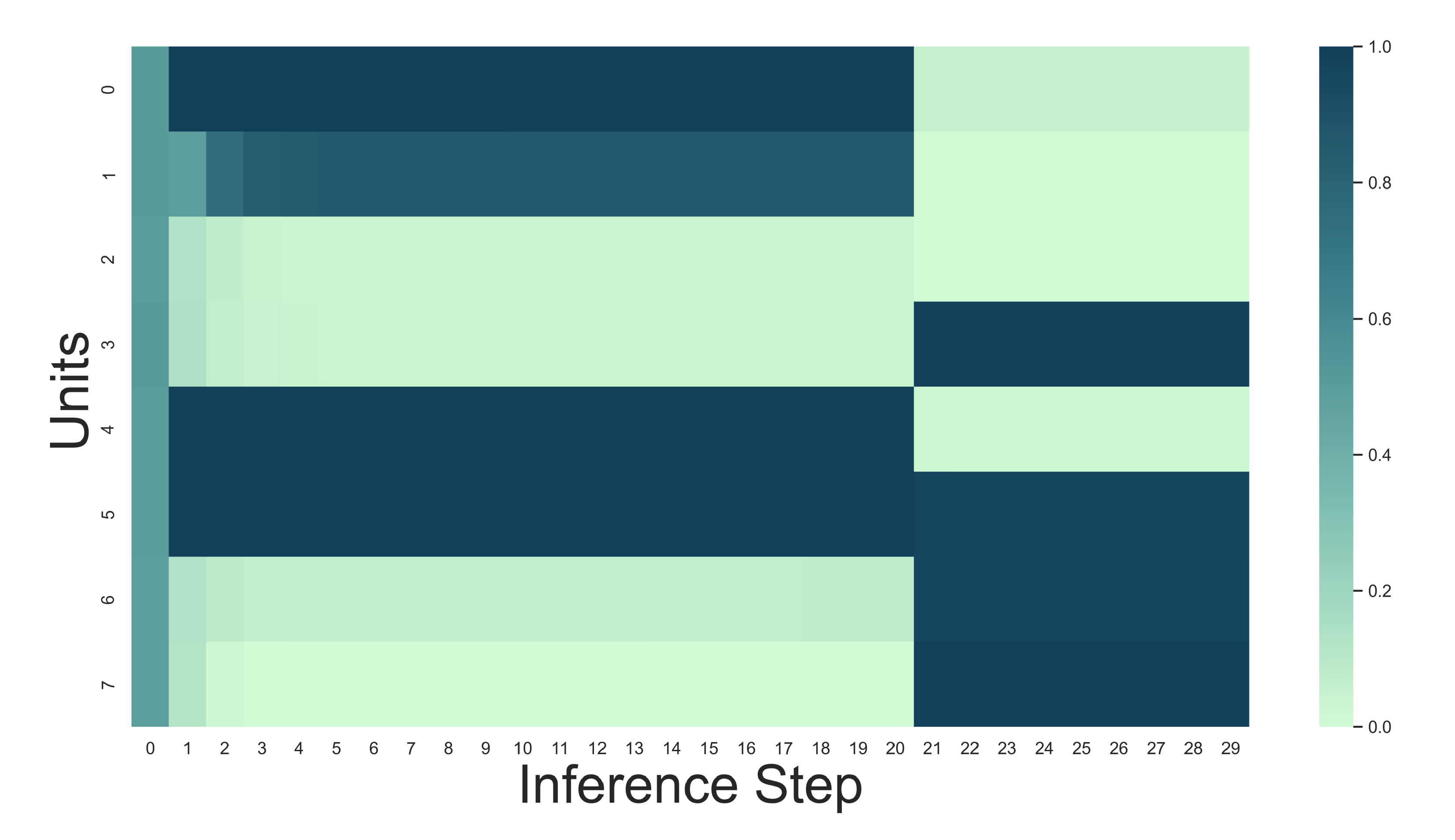}
    \caption{Units Activation of Redundant Information. Dark indicates active and light indicates inactive.}
    \label{fig:redundant}
    \vspace{-15pt}
\end{figure*}
In order to have better generalization, the model needs to correctly identify and process redundant information. To test whether IIU has this stability, we artificially add zero input as redundant information in the process of reasoning. We select the reasoning step as 30 and add redundant information in the last 10 reasoning. As shown in Figure \ref{fig:redundant}, the units group activated when processing key information is 0, 1, 4, 5. After we add redundant information, 3, 5, 6, and 7 are activated. So we can see that IIU can well identify redundant information and process it separately.

\section{Conclusion}
In this paper, we propose Independent Inference Units (IIU) for visual question answering requiring external knowledge. IIU obtains clues hidden in complex multi-modal information by disentangling reasoning information from the functional level to improve the model's performance. In addition, we also propose a Memory Update module to maintain semantic information and help the model resist the influence of noise. IIU as a flexible model achieved a new state-of-the-art result, and surpassing basic pretrained multi-modal models. In the future, we would like to explore whether the dynamic selection of reasoning steps can be realized.

\subsubsection{Acknowledgments.} This work was supported by the Central Guidance for Local Special Project (Grant No. Z231100005923044) and the Climbing Plan Project (Grant No. E3Z0261).
%
%
%
\bibliographystyle{splncs04}
\bibliography{ref}




\end{document}